\documentclass{article}

\PassOptionsToPackage{numbers,sort&compress}{natbib}

\usepackage[preprint]{main}

\usepackage[utf8]{inputenc} % allow utf-8 input
\usepackage[T1]{fontenc}    % use 8-bit T1 fonts
\usepackage{hyperref}       % hyperlinks
\usepackage{url}            % simple URL typesetting
\usepackage{booktabs}       % professional-quality tables
\usepackage{amsfonts}       % blackboard math symbols
\usepackage{nicefrac}       % compact symbols for 1/2, etc.
\usepackage{microtype}      % microtypography
\usepackage{xcolor}         % colors
\usepackage{graphicx}
\usepackage{subcaption}
\usepackage{caption}
\usepackage{amsmath}
\usepackage{flafter}

\usepackage{makecell}
\usepackage{multirow}
\usepackage[table]{xcolor}
\usepackage{colortbl}   % 用于表格底色 \rowcolor
\usepackage{pifont}     % 用于勾号和叉号
\newcommand{\cmark}{\ding{51}} % 勾
\newcommand{\xmark}{\ding{55}} % 叉
\usepackage{algorithm}
\usepackage{algorithmic}

\title{OmniRefine: Alignment-Aware Cooperative Compression for Efficient Omnimodal Large Language Models}

\author{%
  Yuchen Deng$^{1,2}$,
  Zidang Cai$^{1}$,
  Hai-Tao Zheng$^{1,2}$,
  Jie Wang$^{1,2}$,
  Feidiao Yang$^{2}$,
  and Yuxing Han$^{1\dagger}$ \\
  $^{1}$ Tsinghua Shenzhen International Graduate School, Tsinghua University \\
  $^{2}$ Pengcheng Laboratory \\
  \texttt{\{dyc23,jie-wang24\}@mails.tsinghua.edu.cn} \\
  \texttt{\{zheng.haitao,yuxinghan\}@sz.tsinghua.edu.cn} \\
  \texttt{caizidang@gmail.com, yangfd@pcl.ac.cn} \\
  \vspace{0.5em}
  $^{\dagger}$Corresponding authors.
}

\begin{document}

\maketitle

\begin{abstract}
  Omnimodal large language models (Omni-LLMs) show strong capability in audio-video understanding, but their practical deployment remains limited by high inference cost of long video streams and dense audio sequences. Despite recent progress, existing compression methods for Omni-LLMs typically rely on fixed or native compression units, which can disrupt cross-modal correspondence and the complementary information required for audio-video reasoning, making it difficult to improve inference efficiency while stably preserving performance. To address this, we propose OmniRefine, a training-free two-stage framework for efficient audio-visual token compression in Omni-LLMs. First, Correspondence-Preserving Chunk Refinement refines native chunk boundaries into cross-modally aligned compression units through frame-audio similarity and dynamic programming. Second, Modality-Aware Cooperative Compression jointly compresses video and audio tokens within each refined unit to reduce redundancy while preserving critical evidence. Extensive experiments show that OmniRefine achieves a better efficiency-performance trade-off than strong baselines and maintains stable performance under lower compression ratios. On WorldSense, it still reaches 46.7\% accuracy at a 44\% token retention ratio, nearly matching the full-token baseline. The code and interface will be released to facilitate further research.
\end{abstract}

\section{Introduction}
Multimodal Large Language Models (MLLMs)~\cite{wang2024qwen2,bai2025qwen3,li2024llava,liu2024llavanext,cheng2024videollama,liu2024improved,li2023blip,chen2024internvl,sellergren2025medgemma,li2025videochat,lin2024video,deng2025beyond,deng2025avatarsync} have rapidly extended the capabilities of large language models from static image-text understanding to audio-visual perception and reasoning. In particular, building on the progress of video large language models (VLLMs), the rise of omnimodal large language models (Omni-LLMs)~\cite{shu2025audio,xu2025qwen25omnitechnicalreport,xu2025qwen3,team2026qwen3,yang2025humanomniv2,ge2025arc} further enables joint modeling of visual and acoustic inputs.

\begin{figure*}[t]
    \centering
    \includegraphics[width=0.98\textwidth]{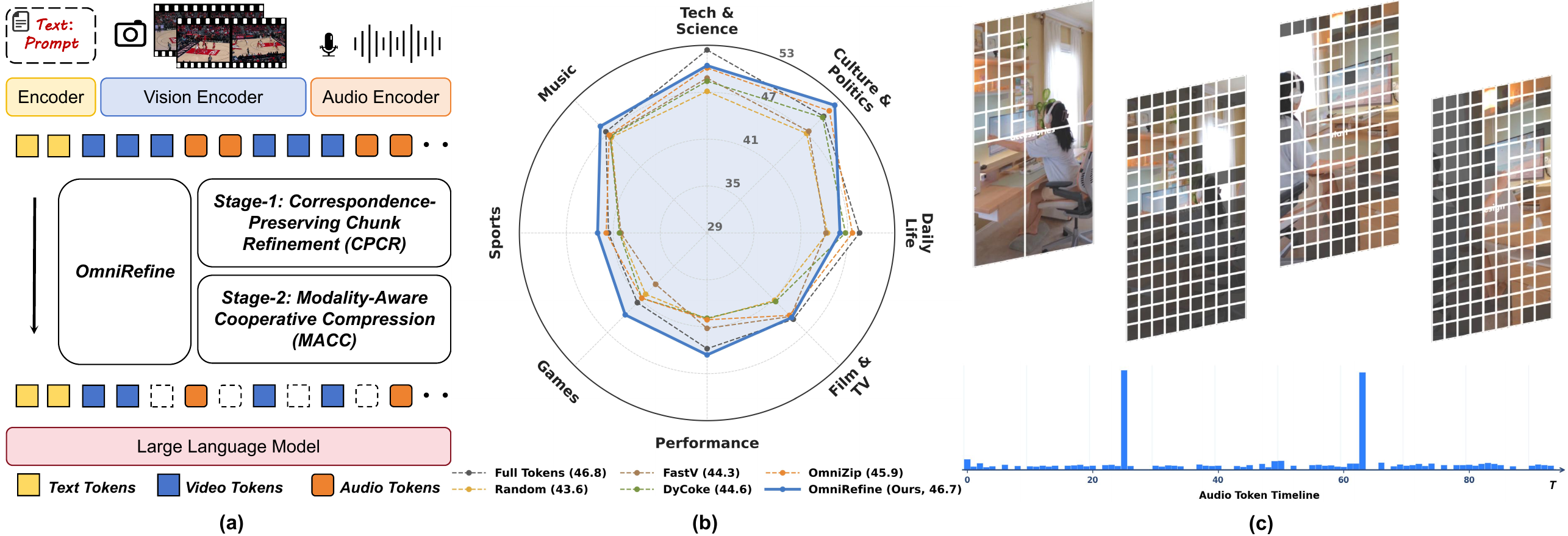}
    \caption{(a) Overview of OmniRefine. (b) OmniRefine outperforms baselines on WorldSense and nearly matches the full-token baseline at a 44\% retention ratio. (c) Visualization of token retention, where gray regions are pruned and the audio timeline marks retained anchor-related positions.}
    \label{fig:overview_1}
    \vspace{-0.5em}
\end{figure*}

However, the deployment of Omni-LLMs still faces severe efficiency bottlenecks~\cite{shen2025fastvid,shao2025holitom,shao2025tokens,hyun2025multi,tao2025omnizip}. This is mainly because long video streams and dense audio token sequences substantially enlarge the inference context, while the quadratic complexity of attention further increases computational and memory overhead. As a result, token compression has emerged as an important direction for accelerating inference. Early token compression methods were developed mainly for VLLMs, where spatial and temporal redundancy is reduced by pruning or merging visual tokens~\cite{chen2024image,huang2025prunevid,shang2025llava,shen2025fastvid,hyun2025multi,tan2025tokencarve,xing2024pyramiddrop,tao2025dycoke,yang2025topv,yang2025visionzip,ye2025fit}. More recent methods have begun to incorporate cross-modal structure into Omni-LLM acceleration, for example through asymmetric compression and cross-modal calibration to better coordinate visual and audio token reduction~\cite{tao2025omnizip,jung2026fastav,jiang2026acckv,gong2025echoingpixels,ding2026omnisift,li2026dash}. Nevertheless, existing compression methods still typically rely on fixed or native compression units, which can easily disrupt cross-modal correspondence and the complementary information required for audio-video reasoning, making it difficult to improve inference efficiency while stably preserving performance.

To investigate this issue, we conduct an empirical analysis of audio-visual correspondence in Qwen2.5-Omni-7B~\cite{xu2025qwen25omnitechnicalreport}. Native temporal chunk boundaries do not always reflect local audio-visual correspondence~\cite{vaswani2017attention}. Some audio tokens may maintain stronger correspondence with video tokens in adjacent chunks. Moreover, within shared compression units, video and audio exhibit different redundancy structures: the former is dominated by spatial and temporal redundancy, whereas the latter relies more on semantic constraints due to the temporal continuity and partial overlap of neighboring audio tokens. This indicates that cross-modally aligned compression units and cooperative compression can better balance the inference efficiency and accuracy of Omni-LLMs.

To this end, we propose \textbf{OmniRefine}, a training-free framework for two-stage audio-visual token compression in Omni-LLMs. As shown in Fig.~\ref{fig:overview_1}(a), OmniRefine first introduces \textbf{Correspondence-Preserving Chunk Refinement (CPCR)}, which uses frame-audio similarity and dynamic programming to refine chunk boundaries into cross-modally aligned compression units. Second, OmniRefine applies \textbf{Modality-Aware Cooperative Compression (MACC)} to exploit cross-modal complementarity, where the video branch reduces spatial and temporal redundancy through a tree-structured strategy, while the audio branch compresses continuous acoustic content under semantic-anchor constraints, with its retention budget adaptively modulated by the video-side retention ratio.

Extensive experiments on WorldSense, AVUT, and VideoMME demonstrate that OmniRefine consistently achieves a better efficiency-performance trade-off than strong baselines. As illustrated in Fig.~\ref{fig:overview_1}, OmniRefine reaches 46.7\% accuracy on Qwen2.5-Omni-7B at a 44\% token retention ratio, nearly matching the full-token baseline. In addition, the visualization further shows that it preserves key audio-video tokens while pruning redundant regions.
Crucially, OmniRefine is training-free and supports KV-cache reuse, making multi-turn reasoning more efficient at lower cost.

In summary, our contributions are listed as follows:
\begin{itemize}
    \item Existing Omni-LLM acceleration methods overlook cross-modal correspondence, making it difficult to balance inference efficiency and quality. Therefore, we propose OmniRefine, a training-free framework for audio-visual token compression.
    \item We further develop a two-stage compression method in which CPCR first refines native chunk boundaries into correspondence-preserving compression units, and MACC then performs modality-aware cooperative compression within each unit.
    \item Extensive experiments on audio-visual benchmarks demonstrate that OmniRefine achieves a more favorable efficiency–performance trade-off than competitive Omni-LLM compression baselines, while remaining training-free and compatible with KV-cache reuse.
\end{itemize}

\section{Related work}
\paragraph{Omnimodal Large Language Models.}
In recent years, multimodal large language models (MLLMs) have evolved from static image-text understanding to reasoning over audio-visual scenarios. Unlike conventional MLLMs~\cite{,bai2025qwen3,cheng2024videollama,chen2024internvl,sellergren2025medgemma,li2025videochat,lin2024video,zhang2024llava,sun2024video}, which process different modalities in a relatively separate manner, Omni-LLMs aim to handle heterogeneous inputs, including text, images, video, and audio, within a unified framework~\cite{xu2025qwen25omnitechnicalreport,xu2025qwen3,team2026qwen3,li2024baichuan,fu2024vita}. Audio-visual understanding has become a key research problem~\cite{ye2025omnivinci,tong2025interactiveomni,ai2025ming,xie2024mini}, as it constitutes a core form of real-world interaction. Along a shared temporal axis, visual and acoustic signals jointly characterize event dynamics: video provides spatial layout, object states, and temporal motion, while audio conveys semantic content, sound events, and temporal cues. Therefore, effective audio-visual understanding requires not only modeling cross-modal temporal relationships, but also fully exploiting the complementary information carried by the two modalities.

\paragraph{Token Compression in Multimodal LLMs.}
Existing token compression methods have been developed mainly for images~\cite{bolya2022token,chen2024image,shang2025llava,tan2025tokencarve,xing2024pyramiddrop,yang2025topv}, videos~\cite{chen2025streamingtom,huang2025prunevid,shao2025holitom,shen2025fastvid,shen2024longvu,tao2025dycoke,hyun2025multi,fu2024framefusion}, and other single-modal inputs~\cite{,dao2023flashattention,dao2022flashattention,shah2024flashattention,lee2025token,li2023accelerating,lin2025speechprune,sun2024video}, primarily exploiting intra-modal redundancy through pruning or merging. However, such methods are not directly suitable for scenarios requiring cross-modal cooperative reasoning, especially when audio and video jointly constitute event evidence along a shared temporal axis~\cite{tao2025omnizip}. This is because single-modal compression can disrupt cross-modal correspondence and consequently degrade downstream reasoning quality.

Recently, to improve the inference efficiency of Omni-LLMs, token compression methods have begun to explicitly incorporate cross-modal structure~\cite{ding2026omnisift,li2026dash,yin2026vllm}. OmniZip~\cite{tao2025omnizip} adopts an asymmetric compression paradigm, using audio to guide dynamic video compression; AccKV~\cite{jiang2026acckv} coordinates the retention of audio and video KV caches through cross-modal calibration; FastAV~\cite{jung2026fastav} employs attention-based two-stage token pruning; and EchoingPixels~\cite{gong2025echoingpixels} performs adaptive compression over joint audio-video tokens under early cross-modal interaction. Despite this progress, existing methods still struggle to maintain stable reasoning accuracy while improving inference efficiency. A key reason is that they overlook cross-modal correspondence and complementary evidence during compression. Differently, OmniRefine refines chunk boundaries to construct cross-modally aligned compression units, and performs cooperative compression by exploiting complementary information from audio and video, thereby achieving a better balance between inference efficiency and accuracy.

\section{Method}
We propose OmniRefine, a training-free audio-visual token compression framework for efficient Omni-LLMs inference. OmniRefine follows a two-stage design that jointly considers cross-modal alignment and modality complementarity. It first refines the native chunks into correspondence-preserving compression units, and then performs cooperative token compression within each unit. The module is applied once to encoded tokens before the LLM prefill stage. Furthermore, our method is question-agnostic, enabling KV-cache reuse across different questions over the same video.

\subsection{Motivating Analysis}
\label{sec:motivating_analysis}
\begin{figure*}[t]
    \centering

    % left column
    \begin{minipage}[t]{0.42\textwidth}
        \vspace{0pt}
        \centering
        \begin{subfigure}[t]{\textwidth}
            \vspace{0pt}
            \centering
            \includegraphics[width=\textwidth]{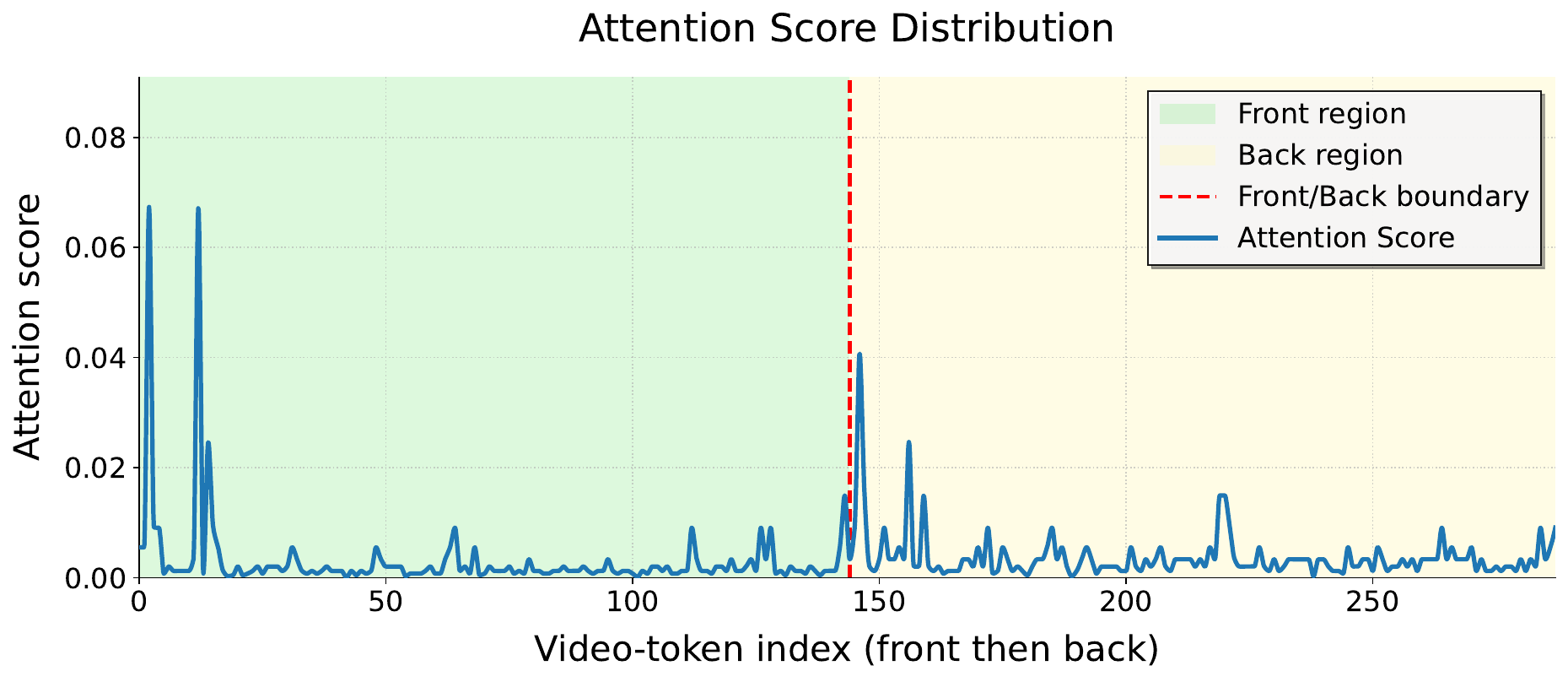}
            \caption{}
            \label{fig:motiv_attn_a}
        \end{subfigure}

        \vspace{0.4em}

        \begin{subfigure}[t]{\textwidth}
            \vspace{0pt}
            \centering
            \includegraphics[width=\textwidth]{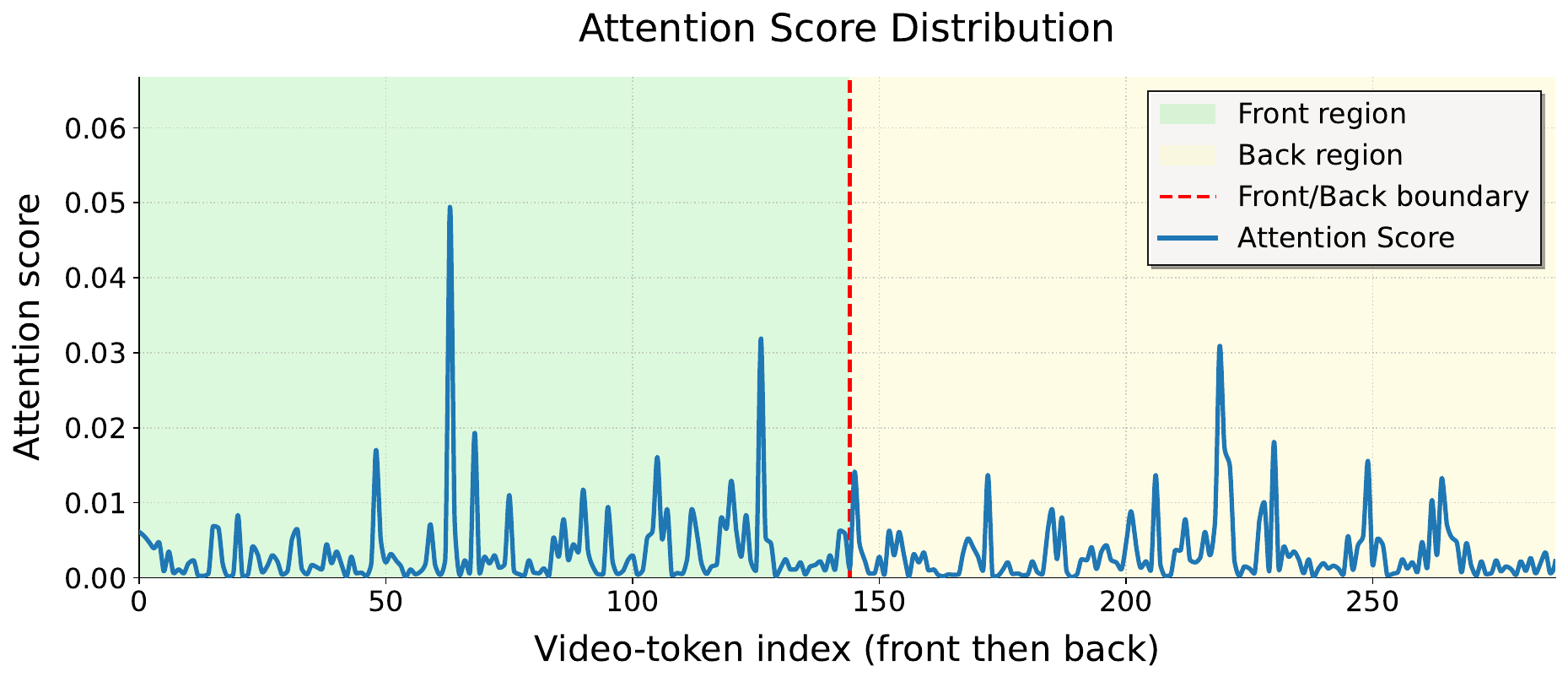}
            \caption{}
            \label{fig:motiv_attn_b}
        \end{subfigure}
    \end{minipage}
    \hfill
    % right column
    \begin{minipage}[t]{0.466\textwidth}
        \vspace{0pt}
        \centering
        \begin{subfigure}[t]{\textwidth}
            \vspace{0pt}
            \centering
            \includegraphics[width=\textwidth]{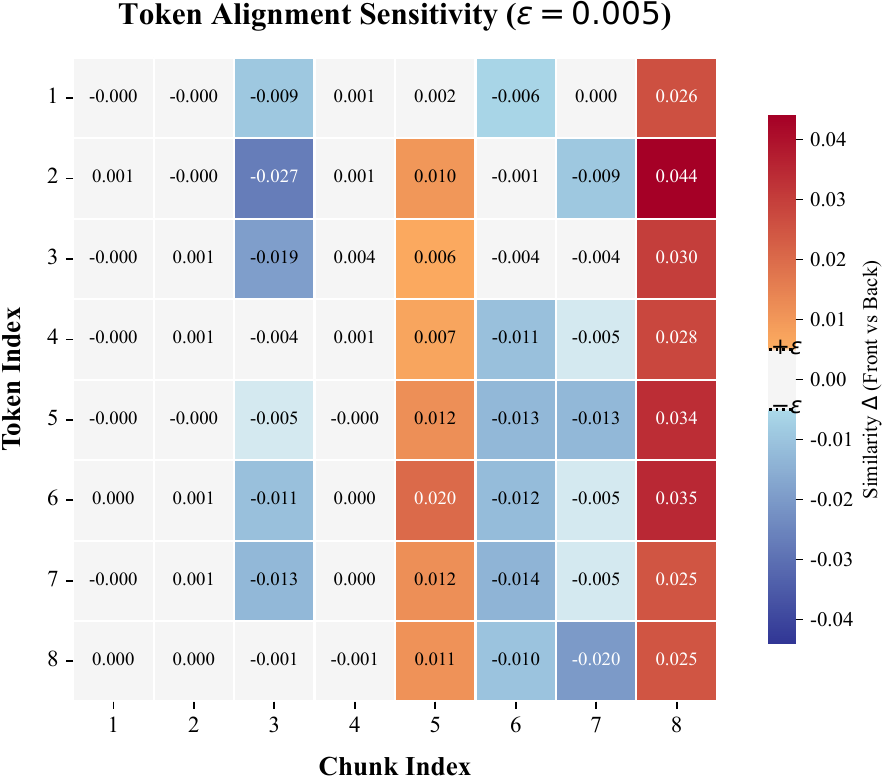}
            \caption{}
            \label{fig:motiv_sim}
        \end{subfigure}
    \end{minipage}

    \caption{Motivating analysis of native chunk boundaries in Qwen2.5-Omni.
    (a) Shallow-layer (layer 0) cross-modal attention of a boundary-adjacent audio token.
    (b) Deeper-layer (layer 8) cross-modal attention of the same token.
    (c) Frame-audio correspondence near native chunk boundaries.}
    \vspace{-0.5em}
    \label{fig:motiv_all}
\end{figure*}

In Qwen2.5-Omni, audio and video tokens are organized into fixed-duration interleaved chunks according to temporal position indices, providing a coarse synchronization prior between modalities. However, these boundaries do not always align with local audio-visual correspondence, so directly using native chunks as compression units may separate locally coherent cross-modal evidence.

To analyze this, we examine the cross-modal attention distributions between audio and video tokens. As shown in Fig.~\ref{fig:motiv_all}(a), although shallow-layer attention is strongly influenced by positional bias, some boundary-adjacent audio tokens in the back chunk already exhibit stronger peak responses to front video tokens. This tendency becomes more pronounced at deeper layers: Fig.~\ref{fig:motiv_all}(b) shows that the same audio token assigns a larger overall attention mass to front video tokens than to back video tokens. This suggests that the cross-modal evidence may extend across native chunk boundaries.
In addition, we use frame-audio cosine similarity in the shared representation space of Omni-LLMs to characterize local audio-visual correspondence. Specifically, video tokens within each frame are aggregated into frame embeddings and compared with boundary-adjacent audio tokens. As shown in Fig.~\ref{fig:motiv_all}(c), some audio tokens on the back side remain closer to the front-frame embedding, indicating that fixed native boundaries may split locally coherent audio-visual evidence.

Taken together, although native chunks provide a coarse temporal synchronization prior, they do not always align with the local audio-visual evidence structure. This motivates OmniRefine to refine the native temporal partition before token compression, and construct correspondence-preserving compression units. However, improved compression units alone are insufficient for efficient audio-visual compression. Under shared chunk boundaries, video and audio still exhibit different redundancy structures: the former depends on spatial layout and temporal dynamics, whereas the latter is dependent on semantic content and temporal continuity. Therefore, cooperative compression is better suited to preserving heterogeneous audio-visual evidence than compression dominated by a single modality. Motivated by this, OmniRefine further adopts a modality-specialized cooperative compression strategy under shared chunk boundaries.

\begin{figure*}[t]
    \centering
    \includegraphics[width=0.96\textwidth]{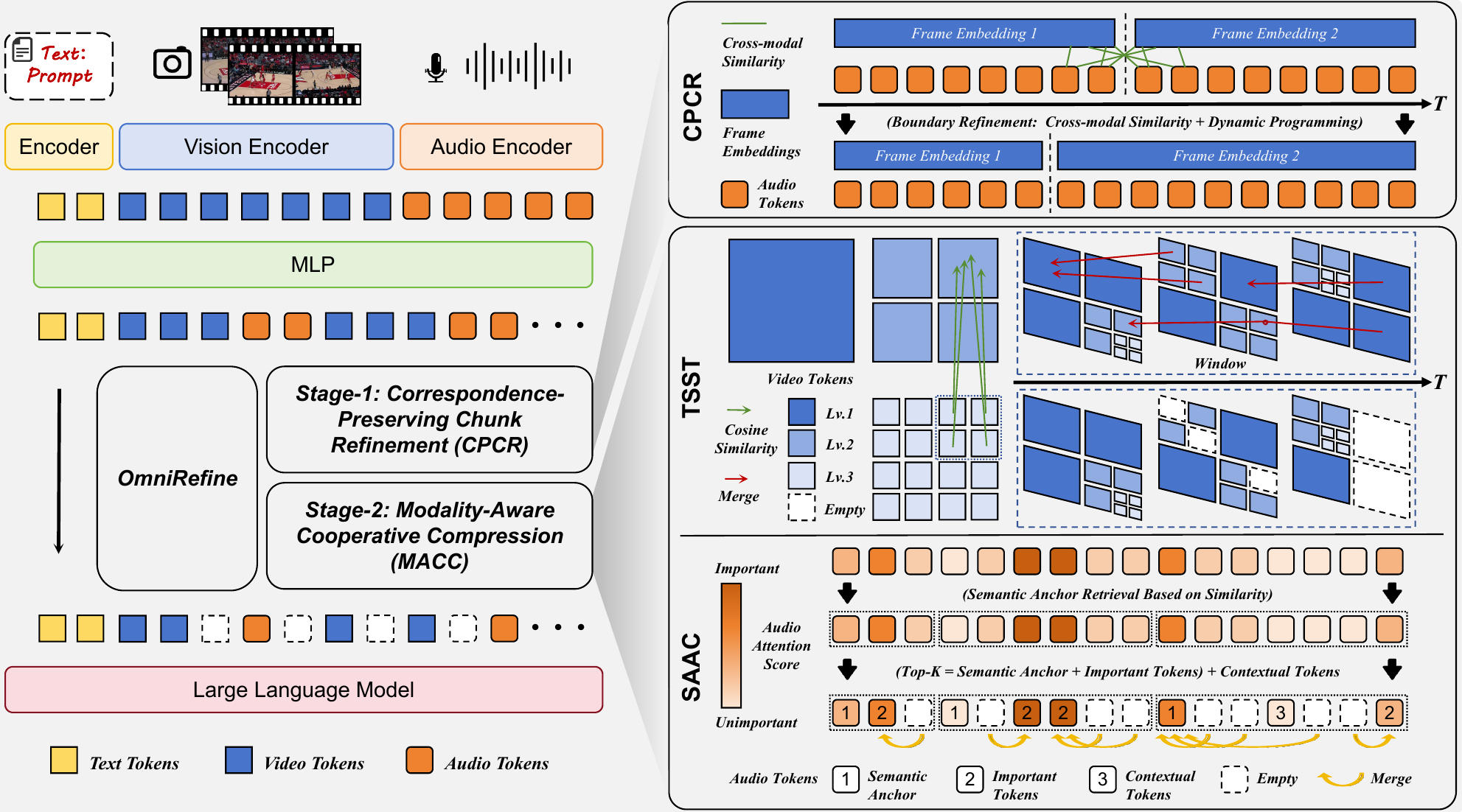}
    \caption{Overview of OmniRefine. Given encoded audio-visual tokens, OmniRefine first applies CPCR to refine native chunks into correspondence-preserving compression units, and then performs MACC within each refined chunk. The compressed tokens are finally passed to the LLM prefill stage.}
    % The compressed audio-visual tokens are finally reassembled and passed to the LLM prefill stage for downstream inference.
    \label{fig:overview}
    \vspace{-0.5em}
\end{figure*}

\subsection{Overview of OmniRefine}
OmniRefine is a training-free framework for two-stage audio-visual token compression in Omni-LLMs. As shown in Fig.~\ref{fig:overview}, it operates on encoded audio and video tokens before the LLM prefill stage. In the first stage, OmniRefine performs Correspondence-Preserving Chunk Refinement (CPCR), which refines native chunk boundaries into cross-modally aligned compression units. Specifically, video tokens within each frame are aggregated into frame-level embeddings and compared with audio token under a native neighborhood constraint. Based on the resulting frame-audio correspondence field, OmniRefine applies constrained dynamic programming to jointly refine video-frame and audio-token boundaries, producing chunks that better match the audio-visual evidence structure.

In the second stage, OmniRefine performs Modality-Aware Cooperative Compression (MACC) within each refined chunk. On the video side, it applies tree-structured spatio-temporal compression (TSST) to reduce spatial and temporal redundancy. On the audio side, it performs semantic-anchor audio compression (SAAC) to preserve semantic continuity while grouping and merging locally related tokens. To enable cross-modal cooperation, the audio branch further adjusts its retention budget according to the video-side retention ratio. Finally, the compressed audio-visual tokens are reassembled in temporal order and passed to the LLM prefill stage for downstream inference.

\subsection{Correspondence-Preserving Chunk Refinement}
Based on the analysis in Sec.~\ref{sec:motivating_analysis}, native temporal chunks are better treated as coarse alignment priors rather than compression units. Thus, we propose Correspondence-Preserving Chunk Refinement (CPCR), which refines chunk boundaries using frame-audio similarity to construct correspondence-preserving compression units. This process is formulated as a temporally constrained joint segmentation problem over video frames and audio tokens, and solved with dynamic programming.
\paragraph{Frame-audio correspondence modeling.}
Let $\{\mathbf{v}_{f,p}\}_{p=1}^{P_f}$ denote the encoded video tokens in frame $f$, and let $\{\mathbf{a}_t\}_{t=1}^{N}$ denote the encoded audio tokens. We compute frame-audio similarity by first aggregating video tokens within each frame into a frame-level representation and then comparing it with each audio token:
\begin{equation}
S_{f,t}=\cos\!\left(\frac{1}{P_f}\sum_{p=1}^{P_f}\mathbf{v}_{f,p},\mathbf{a}_t\right).
\end{equation}

The resulting similarity matrix characterizes frame-audio correspondence. In addition, to preserve temporal priors and confine boundary refinement to native neighborhoods, we only consider valid frame-audio correspondences within those neighborhoods. Let \(c_v(f)\) and \(c_a(t)\) denote the native video bucket of frame \(f\) and the native audio bucket of audio token \(t\). We define a binary mask as
\begin{equation}
M_{f,t}=\mathbb{I}\!\left[c_a(t)\in\mathcal{N}(c_v(f))\right],
\end{equation}
where \(\mathcal{N}(k)\) denotes the valid audio neighborhood of bucket \(k\), consisting of the current bucket and its immediate temporal neighbors, except at the sequence boundaries where only the bucket itself is retained. Therefore, the corresponding masked field is given by
\begin{equation}
\tilde{S}_{f,t}=M_{f,t}S_{f,t},
\end{equation}
which preserves local frame-audio correspondence within native temporal neighborhoods while confining boundary refinement to locally misaligned regions.
\paragraph{Dynamic programming for chunk refinement.}
A candidate refined chunk consists of a contiguous video-frame interval \([i+1,u]\) and a contiguous audio-token interval \([j+1,q]\). We define its score as the average similarity over valid frame-audio pairs within the block:
\begin{equation}
\phi(i,u,j,q)=
\frac{
\sum_{f=i+1}^{u}\sum_{t=j+1}^{q}\tilde{S}_{f,t}
}{
\sum_{f=i+1}^{u}\sum_{t=j+1}^{q}M_{f,t}
}.
\end{equation}
This score measures the internal consistency of a candidate audio-visual block in the shared representation space, where higher values indicate better agreement with the local audio-visual evidence structure.
Based on this score, CPCR formulates chunk refinement as a constrained monotonic segmentation problem over video frames and audio tokens. Let \(D[u,q]\) denote the best segmentation score for the first \(u\) video frames and the first \(q\) audio tokens. The recurrence is
\begin{equation}
D[u,q]=\max_{(i,j)}
\left[
D[i,j]+\phi(i,u,j,q)-\lambda_c
\right],
\label{eq:dp_recursion}
\end{equation}
where \(\lambda_c\) is a chunk regularization term that discourages over-fragmentation. The optimization enforces monotonic progression, continuous coverage, and predefined length constraints in both modalities. Thus, dynamic programming balances local frame-audio correspondence against global segmentation consistency, and traceback yields a sequence of refined audio-visual chunks. These chunks preserve the coarse synchronization structure of the native temporal partition while better matching local audio-visual evidence, and are used as the compression units for the second stage.

\subsection{Modality-Aware Cooperative Compression}
Given the refined chunks produced by CPCR, OmniRefine performs modality-aware cooperative compression within each compression unit. The video branch applies Tree-Structured Spatio-Temporal Compression (TSST), while the audio branch adopts Semantic-Anchor Audio Compression (SAAC), each tailored to the redundancy structure of its modality. Because audio tokens typically encode local acoustic content, neighboring tokens often exhibit temporal continuity and partial overlap. Accordingly, the audio branch references the video-side retention ratio in budget allocation, enabling cross-modal cooperation while preserving semantic continuity.

\paragraph{Tree-Structured Spatio-Temporal Compression.}
For a refined chunk \(g\), let \(\mathcal{V}^{(g)}\) denote its video tokens. Using the original spatial position encoding, we reorganize these tokens into frame-wise 2D grids. As illustrated in Fig.~\ref{fig:overview}, OmniRefine performs spatial compression within each frame through a coarse-to-fine tree-structured search. Specifically, each frame is organized into a multi-scale spatial hierarchy, where a coarse-level node corresponds to a larger 2D region, and each parent node is connected to \(2\times 2\) child regions.
Based on this hierarchy, OmniRefine performs top-down granularity decisions. If all child regions remain sufficiently similar to the parent, the parent node is retained:
\begin{equation}
\cos\bigl(\mathbf{z}(R), \mathbf{z}(R_c)\bigr)\ge \tau_s,\quad \forall R_c \in \mathcal{C}(R)
\end{equation}
where \(\tau_s\) is the spatial similarity threshold and \(\mathcal{C}(R)\) denotes the set of child regions of \(R\). Otherwise, the region is subdivided until finer-grained child regions are reached. This coarse-to-fine search preserves large homogeneous regions while assigning finer-grained representations to complex ones.

After spatial node construction, OmniRefine further performs temporal de-redundancy across adjacent frames. Let \(n_i^{(t-1)}\) and \(n_j^{(t)}\) denote two nodes from consecutive frames whose spatial supports overlap or contain one another. We merge \(n_j^{(t)}\) into the representative of the previous frame when
\begin{equation}
\cos\bigl(\mathbf{z}(n_i^{(t-1)}), \mathbf{z}(n_j^{(t)})\bigr)\ge \tau_t
\end{equation}
where \(\tau_t\) is the temporal similarity threshold. The merged representative is updated by weighted averaging, and only the surviving nodes are mapped back to the visual token mask. The resulting retained set can be written as
\begin{equation}
\mathcal{K}^{(g)}_v=\operatorname{Compress}_v\!\left(\mathcal{V}^{(g)};\tau_s,\tau_t\right).
\end{equation}
In this way, the video branch removes both frame-internal spatial redundancy and cross-frame temporal redundancy while preserving object layout and motion-relevant structure.

\begin{table}[!t]
  \caption{\textbf{Comparison of different methods on WorldSense.} FLOPs calculation considers only multimodal tokens from audio and video inputs. \textbf{Best} result is in bold, \underline{second best} is underlined.}
  \setlength{\tabcolsep}{3.5pt}
  \renewcommand{\arraystretch}{1}
  \label{performance-table}
  \centering
  \footnotesize
  \begin{tabular}{l c c c c c c c c c c c}
    \toprule
    Method & \makecell{Retained\\Ratio} & \makecell{FLOPs\\Ratio} & \makecell{Tech \&\\Science} & \makecell{Culture \&\\Politics} & \makecell{Daily\\Life} & \makecell{Film \&\\TV} & \makecell{Perfor-\\mance} & Games & Sports & Music & Avg. \\
    \midrule
    \multicolumn{12}{c}{Qwen2.5-Omni-7B} \\
    \midrule
    \rowcolor[gray]{.8}Full Tokens & 100\% & 100\% & 52.4 & 50.1 & 48.5 & 44.6 & 43.8 & 41.6 & 41.6 & 47.3 & 46.8 \\
    Random & 55\% & 48\% & 47.1 & 47.0 & 44.4 & 41.2 & 40.0 & 40.1 & 40.1 & 46.3 & 43.6 \\
    FastV & 50\% & 54\% & 48.8 & 47.4 & 44.2 & \underline{44.1} & 41.2 & 38.3 & 40.0 & 46.6 & 44.3 \\
    DyCoke (V\&A) & 50\% & 44\% & 48.4 & 49.9 & 46.7 & 41.4 & 39.9 & 40.8 & 40.2 & 46.5 & 44.6 \\
    OmniZip & 45\% & 39\% & 50.1 & 51.1 & \textbf{47.6} & 43.9 & 40.1 & 40.8 & 41.9 & 46.7 & 45.9 \\
    \rowcolor[gray]{.95}OmniRefine (Ours) & 44\% & 31\% & \textbf{50.4} & \textbf{52.1} & 46.0 & \textbf{44.3} & \textbf{44.6} & \underline{43.8} & \textbf{43.0} & \underline{48.3} & \textbf{46.7} \\
    OmniZip & 35\% & 29\% & 48.3 & 49.5 & \textbf{47.6} & 42.5 & 40.1 & 40.2 & 42.3 & 46.3 & 45.3 \\
    \rowcolor[gray]{.95}OmniRefine (Ours) & 30\% & 20\% & \textbf{50.4} & \underline{51.5} & 45.3 & 43.3 & \underline{44.2} & \textbf{44.6} & \textbf{43.0} & \textbf{48.8} & \underline{46.4} \\
    \midrule
    \multicolumn{12}{c}{Qwen2.5-Omni-3B} \\
    \midrule
    \rowcolor[gray]{.8}Full Tokens & 100\% & 100\% & 51.5 & 50.8 & 45.0 & 45.4 & 43.8 & 42.5 & 44.2 & 46.1 & 46.4 \\
    Random & 55\% & 45\% & 48.2 & 46.3 & 40.7 & 41.4 & 38.6 & 40.0 & 41.8 & 43.4 & 42.8 \\
    FastV & 50\% & 49\% & 50.0 & \textbf{50.5} & 44.1 & 43.0 & 40.5 & 41.6 & 41.8 & 42.1 & 44.4 \\
    DyCoke (V\&A) & 50\% & 40\% & 48.1 & 48.5 & 42.3 & 43.3 & 39.7 & \textbf{43.4} & 42.1 & 43.0 & 44.0 \\
    OmniZip & 45\% & 36\% & \underline{50.1} & \textbf{50.5} & 43.9 & \underline{45.6} & 40.5 & 40.8 & \textbf{43.7} & 43.1 & \underline{45.2} \\
    \rowcolor[gray]{.95}OmniRefine (Ours) & 37\% & 22\% & \textbf{52.2} & 49.5 & \textbf{45.6} & 43.0 & \underline{41.6} & 39.1 & \underline{43.5} & \underline{44.1} & \textbf{45.4} \\
    OmniZip & 35\% & 26\% & 48.8 & 48.9 & 41.8 & \textbf{46.4} & 39.8 & \underline{42.5} & 42.6 & 43.1 & 44.3 \\
    \rowcolor[gray]{.95}OmniRefine (Ours) & 23\% & 18\% & 49.6 & 49.5 & \underline{45.0} & 43.5 & \textbf{41.9} & 39.1 & 40.7 & \textbf{45.1} & 44.7 \\
    \bottomrule
  \end{tabular}
  \vspace{-0.8em}
\end{table}

\paragraph{Semantic-Anchor Audio Compression.}
On the audio side, SAAC combines token saliency, semantic grouping, and cross-modal guidance. For chunk \(g\), let its audio tokens be denoted by \(\mathcal{A}^{(g)}=\{\mathbf{a}^{(g)}_t\}_{t=1}^{N_g}\). First, we identify semantic anchors from adjacent audio-token similarity: whenever the cosine similarity between neighboring tokens falls below a threshold, the latter token is marked as an anchor, partitioning the chunk into local semantic intervals. We then retain a set of dominant audio tokens according to fused attention-based importance scores and keep a small number of contextual anchors from the remaining tokens. Each residual non-anchor token is assigned to its most similar anchor according to local audio similarity, within its semantic interval:
\begin{equation}
\pi(t)=\arg\max_{h\in \mathcal{H}^{(g)}} \cos(\mathbf{a}^{(g)}_t,\mathbf{a}^{(g)}_h).
\end{equation}
where \(\mathcal{H}^{(g)}\) denotes the set of retained anchors in chunk \(g\), and \(\pi(t)\) denotes the anchor assigned to token \(t\). Within each anchor group, we select a small set of merge candidates according to their cross-modal matching scores with the retained video tokens. Let \(r_v^{(g)}\) denote the video-side retention ratio of chunk \(g\). The audio merging ratio is conservatively adjusted according to \(r_v^{(g)}\), with higher visual retention leading to lighter audio compression. The retained audio set is written as
\begin{equation}
\mathcal{K}^{(g)}_a=\operatorname{Compress}_a\!\left(\mathcal{A}^{(g)};\mathcal{H}^{(g)},r_v^{(g)}\right).
\end{equation}
For each anchor \(h\), the selected tokens assigned to it are merged back into the anchor representation through a similarity-weighted update:
\begin{equation}
\tilde{\mathbf{a}}^{(g)}_h=
\frac{
\mathbf{a}^{(g)}_h+\sum_{t\in \mathcal{M}(h)} w_t\,\mathbf{a}^{(g)}_t
}{
1+\sum_{t\in \mathcal{M}(h)} w_t
},
\end{equation}
where \(\mathcal{M}(h)\) denotes the merge set of anchor \(h\), and \(w_t\) is obtained by normalizing the relevance scores within \(\mathcal{M}(h)\). By partitioning semantic intervals, this design preserves key audio tokens while introducing cross-modal guidance through budget allocation and merge weighting. As a result, the audio branch maintains semantic coherence while remaining coordinated with the visual branch.

\begin{table*}[!t]
  \caption{\textbf{Comparison of different token compression methods on AVUTBench and VideoMME.} The FLOPs ratio represents the relative computational overhead compared to the Full Tokens baseline. The ‘-’ symbol indicates that the method (e.g., FastV) failed to execute due to Out-of-Memory (OOM) errors, and such entries are excluded from average calculations.}
  \label{tab:main_comparison}
  \centering
  \footnotesize
  \setlength{\tabcolsep}{5.3pt} 
  \renewcommand{\arraystretch}{1}
  \begin{tabular}{l cc ccccccc cc}
    \toprule
    \multirow{2}{*}{Method} & \multicolumn{2}{c}{Settings} & \multicolumn{7}{c}{AVUTBench} & VideoMME & \multirow{2}{*}{Avg.} \\
    \cmidrule(lr){2-3} \cmidrule(lr){4-10} \cmidrule(lr){11-11}
    & \makecell{Retained} & \makecell{FLOPs} & EL & OR & OM & IE & CC & CM & Avg. & wo & \\
    \midrule
    \multicolumn{12}{c}{\textbf{Qwen2.5-Omni-7B}} \\
    \midrule
    \rowcolor[gray]{.8}Full Tokens & 100\% & 100\% & 38.2 & 67.8 & 59.6 & 85.6 & 44.1 & 66.7 & 64.5 & 66.0 & 100\% \\
    Random & 55\% & 48\% & 38.2 & 64.9 & 55.6 & 80.1 & \textbf{44.7} & 65.0 & 61.0 & 65.4 & 96.9\% \\
    FastV & 50\% & 54\% & 34.1 & 64.3 & \underline{57.1} & 77.6 & 36.4 & 56.4 & 58.4 & - & 90.5\% \\
    DyCoke (V\&A) & 50\% & 44\% & \textbf{38.8} & \underline{67.2} & \textbf{58.2} & 81.9 & 39.0 & 62.4 & 62.0 & 65.5 & 97.7\% \\
    OmniZip & 45\% & 39\% & \underline{38.4} & \underline{67.2} & 56.9 & \underline{85.3} & 42.4 & \textbf{66.0} & \underline{63.0} & \underline{66.3} & \underline{99.1\%} \\
    \rowcolor[gray]{.95} OmniRefine (Ours) & 44\% & 36\% & 36.6 & \textbf{69.2} & 56.2 & \textbf{86.1} & \underline{43.1} & \underline{65.7} & \textbf{63.5} & \textbf{66.4} & \textbf{99.5}\% \\
    \midrule
    \multicolumn{12}{c}{\textbf{Qwen2.5-Omni-3B}} \\
    \midrule
    \rowcolor[gray]{.8}Full Tokens & 100\% & 100\% & 32.9 & 65.3 & 58.4 & 85.0 & 44.1 & 62.6 & 62.2 & 62.6 & 100\% \\
    Random & 55\% & 45\% & 31.7 & 59.2 & 55.4 & 77.3 & \textbf{44.9} & \underline{62.1} & 58.7 & 61.1 & 96.0\% \\
    FastV & 50\% & 49\% & 27.1 & 57.0 & 56.3 & 80.5 & 42.3 & 60.1 & 55.9 & - & 89.9\% \\
    DyCoke (V\&A) & 50\% & 40\% & \underline{31.9} & \underline{64.3} & 57.3 & 82.2 & 40.7 & 61.3 & 60.7 & 61.6 & 98.0\% \\
    OmniZip & 45\% & 36\% & \textbf{32.4} & \textbf{65.0} & \underline{57.7} & \textbf{84.9} & 41.5 & 61.4 & \underline{61.3} & \textbf{62.8} & \underline{99.4\%} \\
    \rowcolor[gray]{.95} OmniRefine (Ours) & 39\% & 28\% & 29.9 & 63.7 & \textbf{58.9} & \underline{84.5} & \underline{44.8} & \textbf{63.0} & \textbf{61.7} & \textbf{62.8} & \textbf{99.8\%} \\
    \bottomrule
  \end{tabular}
  \vspace{-0.8em}
\end{table*}

\section{Experiments}
\subsection{Experimental Setting}
\paragraph{Benchmarks.}
We evaluate our performance on established audio-video understanding benchmarks: WorldSense~\cite{hong2025worldsense}, VideoMME~\cite{fu2025video}, and AVUT~\cite{yang2025audio}. WorldSense assesses the model's ability to jointly understand audio and video across eight distinct domains. VideoMME is widely adopted for video-understanding evaluations, where incorporating audio information can further improve accuracy. AVUT is an audio-centric video understanding benchmark covering six tasks.

\paragraph{Comparison Methods.}
Since token compression methods tailored to omnimodal architectures remain limited, we compare OmniRefine with both the state of the art for Omni-LLMs and representative single-modal baselines. OmniZip~\cite{tao2025omnizip} is the first compression method tailored for Omni-LLMs. FastV~\cite{chen2024image} achieves inference-time, training-free token dropping guided solely by the attention matrix of the L-th layer. DyCoke~\cite{tao2025dycoke} represents the first dynamic token compression strategy proposed for VideoLLMs. In addition, random pruning is included as a control baseline for comparison.
\paragraph{Implementation Details.}
OmniRefine is implemented based on the Qwen2.5-Omni 7B and 3B architectures~\cite{xu2025qwen25omnitechnicalreport}, utilizing NVIDIA L20 (48GB) GPUs. For fair comparison, we adopt the overall FLOPs ratio as the standardized metric. Following prior work, we cap the maximum number of frames at 768 for VideoMME and 128 for WorldSense and AVUT. For hyperparameter settings, OmniRefine uses \(\rho_a=0.3\), \(\rho_v=0.6\), and a contextual ratio of 0.05. In the video branch, the spatial and temporal thresholds are set to \(\tau_s=0.82\) and \(\tau_t=0.58\). In the audio branch, the cross-modal budget coefficient \(\beta\) and the semantic-anchor similarity threshold are both set to 0.4. For CPCR, we use a regularization term \(\lambda_c=0.02\). Additional details are provided in Appendix~\ref{app:Hyperparameter} and~\ref{app:Algorithmic}.

\subsection{Main Results}
We evaluate our proposed OmniRefine on the Qwen2.5-Omni model at two parameter scales (7B and 3B) across three major benchmarks: WorldSense, VideoMME, and AVUT. Specifically, we utilize the LMMs-Eval framework~\cite{zhang2025lmms} for the VideoMME evaluation, while applying a unified testing codebase across all experimental settings for the remaining benchmarks. For comparison, we establish baselines using random pruning, FastV, DyCoke, and OmniZip. Furthermore, OmniRefine is evaluated under various token retention rates to comprehensively analyze the trade-off between model performance and inference overhead. To facilitate a more thorough horizontal comparison, the FLOPs presented in Tab.\ref{performance-table} and Tab.\ref{tab:main_comparison} are normalized into percentages, where the computational cost of the uncompressed full-token baseline is designated as 100\%.

\paragraph{Comparison with State-of-the-Art Methods.}
As shown in Tab.~\ref{performance-table}, OmniRefine consistently performs strongly across diverse audio-video understanding tasks, remaining stable even under aggressive compression. On the 7B model, utilizing a 44\% token retention ratio, OmniRefine achieves 46.7\% accuracy, essentially matching the uncompressed full-token baseline while reducing computational FLOPs by 69\%. When the retention ratio is further reduced to 30\%, OmniRefine exhibits only a minor performance drop and still achieves 46.4\% accuracy, outperforming OmniZip at higher 45\% / 35\% retention budgets (45.9\% / 45.3\%) as well as DyCoke at 50\% retention (44.6\%).

Furthermore, as shown in Tab.~\ref{tab:main_comparison}, OmniRefine maintains strong performance on both AVUT and VideoMME under substantially reduced computational budgets. Across model scales, it overall outperforms OmniZip on AVUT while matching or slightly exceeding the full-token baseline on VideoMME. Even with a 72\% FLOPs reduction, OmniRefine still retains 99.8\% average normalized accuracy, indicating that the proposed design generalizes well across diverse audio-video benchmarks.

\begin{figure*}[t]
  \centering
  % --- 左侧：表格 ---
  \begin{minipage}[c]{0.54\textwidth}
    \centering
    \captionsetup{font=normalsize}
    \captionof{table}{\textbf{Efficiency comparison on the WorldSense benchmark.} We report peak GPU memory usage and inference latency for Qwen2.5-Omni-7B. }
    \label{tab:efficiency-table}
    \footnotesize
    \setlength{\tabcolsep}{2.5pt} % 稍微微调以适应新布局
    \renewcommand{\arraystretch}{1}
    \begin{tabular}{l cccc}
      \toprule
      Method & \makecell[c]{GPU \\ Mem. $\downarrow$} & \makecell[c]{Prefilling \\ Time $\downarrow$} & \makecell[c]{Acc. \\ $\uparrow$} & \makecell[c]{Latency per \\ Example $\downarrow$} \\
      \midrule
      \rowcolor[gray]{.8} Full Tokens & 44G & 2371ms (1.00$\times$) & 46.8 & 10.99s (1.00$\times$) \\
      FastV & \multicolumn{4}{c}{OOM} \\ 
      DyCoke (V\&A) & 36G & 1386ms (1.71$\times$) & 44.6 & 8.59s (1.28$\times$) \\
      OmniZip (45\%) & 32G & 894ms (2.65$\times$) & 45.9 & \underline{7.99s (1.38$\times$)} \\
      OmniZip (35\%) & \underline{30G} & \underline{649ms (3.65$\times$)} & 45.3 & \textbf{7.46s (1.47$\times$)} \\
      \rowcolor[gray]{.95} Ours (30\%) & \textbf{29G} & \textbf{451ms (5.26$\times$)} & 46.4 & 9.59s (1.15$\times$) \\
      \bottomrule
    \end{tabular}
  \end{minipage}
  \hfill % 在两个 minipage 之间填充空白
  % --- 右侧：图片 ---
  \begin{minipage}[c]{0.43\textwidth}
    \centering
    \includegraphics[width=\textwidth]{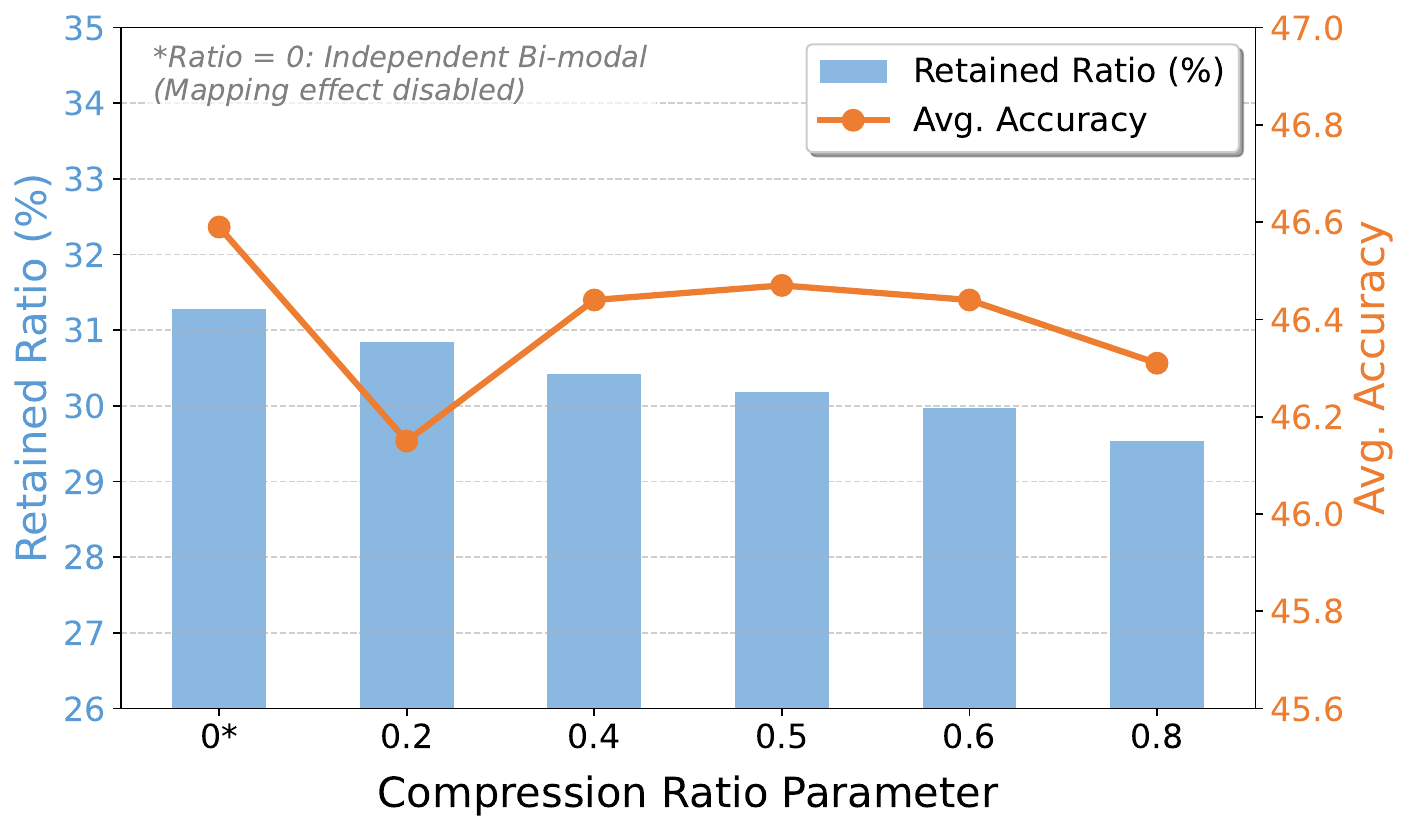}
    \captionsetup{font=normalsize}
    \captionof{figure}{\textbf{Ablation on the audio budget.} Performance across  budget parameters.}
    \label{fig:audio_budget}
  \end{minipage}
\end{figure*}

\begin{figure*}[t]
  \centering
  
  % --- 左侧：表格 (Ablation of Core Modules) ---
  \begin{minipage}[c]{0.49\textwidth}
    \centering
    % \captionof{table}{\textbf{Ablation of OmniRefine's core modules.} We substantiate OmniRefine's design by systematically evaluating the impact of CPCR and MACC. This compares performance variations when omitting these crucial components.}
    \captionsetup{font=normalsize}
    \captionof{table}{\textbf{Ablation of CPCR and MACC in OmniRefine.} Specifically, w/o CPCR uses native chunks with MACC, w/o MACC uses CPCR with OmniZip-style compression, and w/o Both uses native chunks with OmniZip-style compression.}
    \label{tab:ablation_core}
    \footnotesize
    \setlength{\tabcolsep}{3.5pt} % 调整列间距以适应5列布局
    \renewcommand{\arraystretch}{1}
    \begin{tabular}{l c c c c}
      \toprule
      Settings & CPCR & MACC & \makecell[c]{Retained} & WorldSense \\
      \midrule
      \rowcolor[gray]{.95}Full OmniRefine & \cmark & \cmark & 44 & 46.7 \\
      w/o CPCR & \xmark & \cmark & 45 & 46.4 \\
      w/o MACC & \cmark & \xmark & 45 & 46.2 \\
      w/o Both & \xmark & \xmark & 45 & 45.9 \\
      \bottomrule
    \end{tabular}
  \end{minipage}
  \hfill % 分隔左右两个部分
  % --- 右侧：图片 (Ablation on Audio Budget) ---
  \begin{minipage}[c]{0.48\textwidth}
    \centering
    \includegraphics[width=\textwidth]{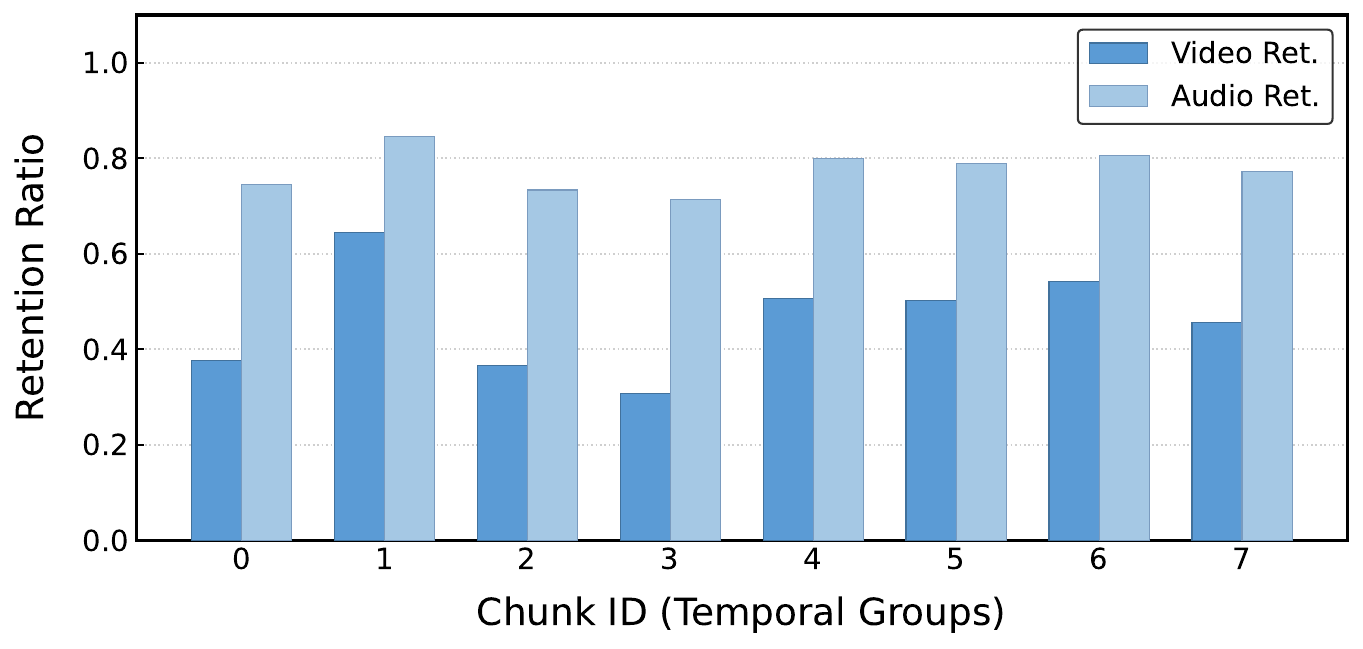}
    \captionsetup{font=normalsize}
    \captionof{figure}{\textbf{Visualization of Dynamic Pruning.} Video and audio retention ratios per chunk.}
    \label{fig:pruning}
  \end{minipage}
  
\end{figure*}

\paragraph{Efficiency Analyses.}
We evaluate the inference latency and memory consumption across four benchmarks. As detailed in Tab.~\ref{tab:efficiency-table}, a comprehensive analysis is conducted on the WorldSense benchmark. The results demonstrate that, on the 7B model, our method achieves a 1.15× speedup in overall inference and a remarkable 5.26× acceleration during the prefilling stage compared to the full-token baseline. Furthermore, our approach substantially mitigates memory overhead. By saving 15GB of GPU memory while retaining approximately 99\% of the original accuracy, this method provides crucial efficiency gains for the practical deployment of OmniLLMs. Since OmniRefine is training-free and supports KV-cache reuse, it is attractive for low-cost multi-turn inference.

\subsection{Ablation Study}
\paragraph{Ablation of CPCR and MACC.}
Tab.~\ref{tab:ablation_core} studies the contributions of CPCR and MACC in OmniRefine. The full model achieves the highest WorldSense score (46.7\%) with the lowest retained ratio (44\%). Removing CPCR or MACC drops performance to 46.4\% and 46.2\% respectively, while increasing the ratio to 45\%. Disabling both yields the lowest accuracy (45.9\%), indicating that CPCR and MACC are complementary for balancing performance and efficiency.

\paragraph{Analysis of audio budget coordination.}
Fig.~\ref{fig:audio_budget} evaluates our audio budget strategy. While independent bi-modal compression (Ratio=0) yields the highest accuracy (46.59\%), it suffers from the highest retained ratio (31.28\%). Our video-referenced modulation achieves an optimal trade-off at Ratio=0.5, maintaining a highly competitive 46.47\% accuracy while reducing the retained ratio to 30.19\%, demonstrating effective adaptive budget coordination. Fig.~\ref{fig:pruning} further visualizes dynamic pruning. The retained ratios of video and audio tokens vary substantially across chunks, showing that OmniRefine allocates modality-specific budgets adaptively rather than applying uniform compression.

\section{Conclusion}
In this paper, we propose OmniRefine, a training-free framework for two-stage audio-visual token compression in Omni-LLMs. It first refines native temporal chunks into correspondence-preserving compression units, and then performs modality-aware cooperative compression within each refined chunk. Notably, under matched compression budgets, OmniRefine preserves model accuracy substantially better than existing baselines, achieving a more favorable trade-off between efficiency and performance. Since OmniRefine is training-free and question-agnostic, it is naturally compatible with efficient inference settings such as KV-cache reuse. While OmniRefine achieves strong compression performance, its reliance on manual hyperparameters is a limitation further discussed in Appendix \ref{sec:limitations}. Future work will explore adaptive hyperparameter selection under different token budgets, enabling chunk refinement and cooperative compression to adjust automatically to input complexity.

\bibliographystyle{plainnat}
\bibliography{references}

\clearpage
\appendix

\section{Hyperparameter Settings}
\label{app:Hyperparameter}

To supplement the configuration details provided in the main text, Table~\ref{tab:hyperparameters} summarizes the comprehensive hyperparameter settings utilized in the OmniRefine framework. While the primary thresholds ($\rho$, $\tau$, $\beta$, $\lambda_c$) control the core semantic filtering behavior, we introduce specific boundary constraints and hardware-aware optimizations to ensure stable feature compression and alignment efficiency.

\textbf{Cross-Modal Budget Bounds.} To prevent extreme pruning that could lead to catastrophic semantic loss, or insufficient compression that undermines acceleration, we define hard boundaries for the token retention mechanisms. The video token retention ratio is strictly bounded within $[0.18, 0.55]$, ensuring that at least $18\%$ of the most critical visual cues are preserved regardless of the scene's sparsity. The audio retention bounds are set to $[0.1, 0.9]$. Additionally, a video budget modulation factor $\alpha=0.15$ is employed to fine-tune the cross-chunk visual budget allocation. 

\textbf{Chunking and DP Alignment Constraints.} For the multimodal alignment module, we impose physical chunking limits to maintain intra-chunk temporal coherence. Specifically, The minimum and maximum sizes of an audio segment are set to 90 and 140 tokens, and each video chunk is constrained to contain between 3 and 5 frames. To accelerate the Dynamic Programming (DP) based joint chunking process, we apply a local alignment window strategy. The DP band ratio is set to 2.0 with a local minimum window size of 48 tokens, effectively reducing the computational overhead of solving the optimal global alignment path without sacrificing matching accuracy.

\begin{table}[h]
\centering
\caption{Detailed hyperparameter configurations for OmniRefine.}
\label{tab:hyperparameters}
\resizebox{0.95\columnwidth}{!}{
\begin{tabular}{lccl}
\toprule
\textbf{Category} & \textbf{Parameter} & \textbf{Value} & \textbf{Description} \\ \midrule
\multirow{3}{*}{\textbf{Main Settings (Main Text)}} 
& $\rho_a, \rho_v$ & 0.3, 0.6 & Global compression ratios \\
& $\tau_s, \tau_t$ & 0.82, 0.58 & Spatial and temporal STTM thresholds \\
& $\beta$ & 0.5 & Audio-to-video cross-modal budget coefficient \\
& $\lambda_c$ & 0.02 & CPCR regularization penalty term \\ \midrule
\multirow{4}{*}{\textbf{Budget Constraints}} 
& $G$ & 3 & Group block size for local token processing \\
& $[v_{\min}, v_{\max}]$ & $[0.18, 0.55]$ & Hard lower/upper bounds for video token retention \\
& $[a_{\min}, a_{\max}]$ & $[0.1, 0.9]$ & Hard lower/upper bounds for audio token retention \\
& $\alpha$ & 0.15 & Video budget modulation factor \\ \midrule
\multirow{4}{*}{\textbf{Chunking \& DP}} 
& $S_{A-\min}$ & 90 & Minimum allowable audio chunk size (tokens) \\
& $S_{A-\max}$ & 140 & Maximum allowable audio chunk size (tokens) \\
& $S_{V-\min}$ & 3 & Minimum allowable video chunk size (frames) \\
& $S_{V-\max}$ & 5 & Maximum allowable video chunk size (frames) \\
& DP Band Ratio & 2.0 & Search bandwidth ratio for DP alignment efficiency \\
& Min DP Window & 48 & Minimum local matching window for DP (tokens) \\
\bottomrule
\end{tabular}
}
\end{table}

\section{Algorithmic Details \& Pseudo-code}
\label{app:Algorithmic}

To address the implementation specifics of the multimodal fusion process, we detail the representation computation, cross-modal budget modulation, and the constrained dynamic programming optimization below.

\subsection{MACC Representation and Audio Budget Allocation}
To clarify the region/node representations $\mathbf{z}(\cdot)$ used in the MACC module, $\mathbf{z}(\cdot)$ is computed as the average-pooled representation of the hidden states of the tokens within the corresponding spatial-temporal region. Specifically, for a set of token embeddings $\{\mathbf{h}_1, \mathbf{h}_2, \dots, \mathbf{h}_k\}$ in a given block, the aggregated representation is defined as:
\begin{equation}
    \mathbf{z} = \frac{1}{k} \sum_{i=1}^k \mathbf{h}_i
\end{equation}

Furthermore, to address the cross-modal budget allocation (illustrated as the ``Compression Ratio Parameter'' in Fig. 4 of the main text), we denote $\rho_a,\rho_v$ as the global base \emph{merging ratios} (compression ratios). Let $m_a$ be the audio merging ratio and $R_v$ be the observed video retention ratio in the current chunk. The video-referenced audio budget is updated by:
\begin{equation}
    m_a = \min \Big( a_{\max}, \max \big(a_{\min}, \, \rho_a - \beta \cdot (R_v - (1-\rho_v)) \big) \Big)
\end{equation}
and the corresponding audio retention ratio is
\begin{equation}
    R_a = 1 - m_a.
\end{equation}
where $(1-\rho_v)$ is the base video retention level implied by $\rho_v$, and $[a_{\min}, a_{\max}]$ are safety bounds on the audio merging ratio to avoid over-pruning or under-compression.

\subsection{CPCR Constrained Dynamic Programming}
A naive global solution for the joint chunking alignment (Eq.~\ref{eq:dp_recursion} in the main text) would require $O(F^2 N^2)$ time complexity, making it computationally prohibitive for long contexts. To mitigate this, we introduce the CPCR Constrained Dynamic Programming algorithm. 

We constrain the search space using physical chunk boundaries
($S_{V-\min}, S_{V-\max}, S_{A-\min}, S_{A-\max}$) together with a Neighborhood Mask $\mathcal{M}$.
The admissible region is further restricted by a DP band ratio ($B$) and a local minimum window size ($W$),
which keeps valid alignments near the temporal diagonal while preserving feasible chunk transitions.
The detailed procedure is given in Algorithm~\ref{alg:cpcr_dp}.

\begin{algorithm}[ht]
\small
\caption{CPCR Constrained Dynamic Programming for Joint Chunking}
\label{alg:cpcr_dp}
\begin{algorithmic}[1]
\REQUIRE Video frame sequence $\mathbf{V}$ (length $F$), audio token sequence $\mathbf{A}$ (length $N$)
\REQUIRE Chunk bounds $S_{V-\min}=3$, $S_{V-\max}=5$, $S_{A-\min}=90$, $S_{A-\max}=140$
\REQUIRE DP band ratio $B=2.0$, minimum DP window $W=48$, penalty $\lambda_c=0.02$
\ENSURE Optimal alignment path $\mathcal{P}^*$

\STATE Initialize DP matrix $D \in \mathbb{R}^{(F+1)\times(N+1)}$ with $+\infty$
\STATE Initialize backpointer matrix $\Pi$ of size $(F+1)\times(N+1)$
\STATE $D[0,0] \gets 0$

\STATE Define expected diagonal position $\hat{j}(i) \gets i\cdot \frac{N}{F}$
\STATE Define admissible mask
\STATE $\mathcal{M}(i,j)=\mathbf{1}\!\left(\left|j-\hat{j}(i)\right|\le \max\!\left(W,\;B\cdot \frac{N}{F}\cdot S_{V-\max}\right)\right)$

\FOR{$i=1$ \TO $F$}
  \FOR{$j=1$ \TO $N$}
    \IF{\textbf{not} isValid$(\mathcal{M}(i,j))$}
      \STATE \textbf{continue} \COMMENT{Prune out-of-band states}
    \ENDIF

    \STATE $D[i,j]\gets +\infty$

    \STATE $i_0^{\min}\gets \max(0,\;i-S_{V-\max})$
    \STATE $i_0^{\max}\gets i-S_{V-\min}$
    \STATE $j_0^{\min}\gets \max(0,\;j-S_{A-\max})$
    \STATE $j_0^{\max}\gets j-S_{A-\min}$

    \FOR{$prev_i=i_0^{\max}$ \TO $i_0^{\min}$}
      \FOR{$prev_j=j_0^{\max}$ \TO $j_0^{\min}$}
        \STATE Compute match score $S_{\text{match}}$ between $\mathbf{V}[prev_i:i]$ and $\mathbf{A}[prev_j:j]$
        \STATE $cost \gets -S_{\text{match}} + \lambda_c \cdot \mathrm{ChunkVariance}$
        \IF{$D[prev_i,prev_j] + cost < D[i,j]$}
          \STATE $D[i,j] \gets D[prev_i,prev_j] + cost$
          \STATE $\Pi[i,j] \gets (prev_i,prev_j)$
        \ENDIF
      \ENDFOR
    \ENDFOR
  \ENDFOR
\ENDFOR

\STATE Backtrack from $\Pi[F,N]$ to $(0,0)$ to obtain $\mathcal{P}^*$
\RETURN $\mathcal{P}^*$
\end{algorithmic}
\end{algorithm}

\textbf{Complexity Reduction:}
By jointly enforcing (i) diagonal neighborhood constraints via $\mathcal{M}$, (ii) local window/band limits, and (iii) chunk-size bounds
$\left[S_{V-\min},S_{V-\max}\right]$ and $\left[S_{A-\min},S_{A-\max}\right]$, we prune a large portion of infeasible transitions compared with the unconstrained global search. Therefore, while the naive formulation scales as $O(F^2N^2)$, the practical runtime of CPCR is governed by the admissible band/window width and the chunk-bound candidate ranges, yielding a substantially smaller effective search space for deployment.

\section{Evaluation Protocol}
\label{app:Evaluation}

\subsection{Hardware \& Latency Profiling}
All inference speed and memory profiling were conducted on a single NVIDIA L20 (48GB) GPU to ensure hardware consistency. To address concerns regarding end-to-end speedups, we strictly decouple the runtime into three distinct stages: preprocessing (CPCR and MACC), prefill, and decoding. In our implementation, preprocessing entails the CPU-side multimodal preparation prior to generation, while prefill latency is measured by timing the first forward pass. End-to-end latency covers the full duration from preprocessing to the completion of generation.

Importantly, the preprocessing overhead of OmniRefine is extremely lightweight—typically taking only a few milliseconds—which is entirely negligible compared to the prefill and decoding stages. The observed gap between the massive prefill acceleration and the more modest end-to-end speedup is an expected behavior of LLMs, as the autoregressive decoding phase dominates the total wall-clock time for long textual outputs. OmniRefine effectively eliminates the computational bottleneck of processing long multimodal contexts without altering the inherent decoding speed.

\subsection{Multi-Turn KV-Cache Simulation}
A major practical advantage of our question-agnostic, training-free compression is its inherent compatibility with KV-cache reuse in multi-turn interactions. Under our protocol, the multimodal context is compressed only once during the initial prefill stage. For subsequent queries ($q_2, q_3, \dots, q_k$) regarding the same media, the refined KV-cache is directly reused without re-triggering the visual/audio encoders or the compression modules. 

This efficiently amortizes the initial context-processing cost over multiple turns. While our primary quantitative evaluation focuses on standard single-turn benchmarks, this simulation protocol theoretically characterizes how OmniRefine yields substantial cumulative efficiency gains and deployment benefits in real-world conversational scenarios.

\subsection{Constant-Budget Alignment \& Reliability}
To ensure ablation fairness and avoid confounding factors from varying compression budgets, our extended ablation studies follow a strict constant-budget protocol. Specifically, we fine-tune the global compression parameters ($\rho_a, \rho_v$) and adjust corresponding constraints to lock the overall token compression rate to a shared target budget ($\rho_{\mathrm{ret}}^{\star}$) across all compared variants. This ensures that the ablation comparisons are directly attributable to architectural contributions (i.e., CPCR and MACC) under matched computational FLOPs. 

Furthermore, to mitigate potential hardware variance and decoding randomness, the latency profiling and key performance metrics are averaged over multiple independent evaluation runs. This standardizes the reported FLOPs ratios, latency, and memory metrics, ensuring the statistical reliability and reproducibility of our results across different models and settings.

\section{Limitations and Future Work}
\label{sec:limitations}

While OmniRefine demonstrates state-of-the-art compression efficiency and robust cross-modal alignment across diverse benchmarks, we identify two primary limitations that present promising avenues for future research.

\textbf{Audio-Dominant and Off-Screen Scenarios.} 
A core design of our Modality-Aware Cooperative Compression (MACC) module is the video-referenced audio budget allocation, which couples audio retention to the visual compression rate. While highly effective for general multimodal scenes, this mechanism may exhibit vulnerabilities in extreme audio-dominant or visually sparse scenarios (e.g., speech over static presentation slides, or critical off-screen sound events). In such cases, a low visual token retention rate could inadvertently lead to the over-compression of information-dense audio tokens. Although we currently mitigate this risk by enforcing hard minimum retention bounds (fail-safes such as $a_{\min}$), future work could introduce dynamic, cross-modal entropy-based fail-safe mechanisms. This would allow the framework to automatically decouple modality budgets when the semantic density of the audio stream significantly outweighs that of the visual stream.

\textbf{Auto-Tuning of Hyperparameters.} 
A secondary limitation lies in the reliance on predefined empirical thresholds (e.g., spatial-temporal thresholds $\tau_s, \tau_t$ and the CPCR regularization penalty $\lambda_c$). Although our extensive evaluations confirm that these default parameters generalize remarkably well across different model scales (7B/3B) and datasets without task-specific modifications, the framework is not entirely parameter-free. To further enhance its plug-and-play capability, future iterations of OmniRefine could leverage automated hyperparameter search (Auto-tuning) or reinforcement learning strategies. This would enable the model to adaptively find the optimal threshold configurations conditioned on specific hardware FLOP constraints and shifting data distributions.

\end{document}